\documentclass{article}

\usepackage[preprint,nonatbib]{neurips_2024}
\usepackage[utf8]{inputenc} 
\usepackage[T1]{fontenc}    
\usepackage{hyperref}       
\usepackage{url}            
\usepackage{booktabs}       
\usepackage{amsfonts}       
\usepackage{nicefrac}       
\usepackage{microtype}      
\usepackage{xcolor}         
\usepackage{amsmath}        
\usepackage{graphicx}       
\usepackage{multirow}       
\usepackage{enumitem}       
\usepackage{lineno}         
\usepackage[export]{adjustbox}  
\usepackage[normalem]{ulem}
\usepackage{amssymb}        
\usepackage{pifont}         
\usepackage{tabularx}       
\usepackage{subcaption}
\usepackage{float}          
\usepackage{footmisc} 

\usepackage{algorithm}
\usepackage{algpseudocode} 
\usepackage{tikz}
\usetikzlibrary{shapes, arrows, positioning}
\usepackage{subcaption}
\usepackage{textcomp}
\usepackage{pdfpages}
\usepackage{pifont}
\usepackage[style=numeric]{biblatex}  
\newcommand{\cmark}{\ding{51}}%
\newcommand{\xmark}{\ding{55}}%
\useunder{\uline}{\ul}{}
\tikzstyle{block} = [rectangle, draw, text width=2.5cm, text centered, rounded corners, minimum height=1.2cm]
\tikzstyle{line} = [draw, -latex']
\DeclareMathOperator*{\argmin}{arg\,min}

\title{CONSULT: Contrastive Self-Supervised Learning for Few-shot Tumor Detection}

\author{
  Sin Chee Chin  \\
  Shenzhen International Graduate School\\
  Tsinghua University \\
  ShenZhen, China \\
  \texttt{chenxz22@mails.tsinghua.edu.cn} \\
  \And
  Xuan Zhang \\
  Shenzhen International Graduate School\\
  Tsinghua University\\
  ShenZhen, China \\
  \texttt{zhangxua22@mails.tsinghua.edu.cn} \\
  \And
  Lee Yeong Khang \\
  ViTrox Technologies Sdn. Bhd.\\
  Penang, Malaysia \\
  \texttt{yeong-khang.lee@vitrox.com} \\
  \And
  Wenming Yang\thanks{Correspondance Author}  \\
  Shenzhen International Graduate School\\
  Tsinghua University \\
  ShenZhen, China \\
  \texttt{yang.wenming@sz.tsinghua.edu.cn} \\
}

\begin{document}

\maketitle

\begin{abstract}

Artificial intelligence aids in brain tumor detection via MRI scans, enhancing the accuracy and reducing the workload of medical professionals. However, in scenarios with extremely limited medical images, traditional deep learning approaches tend to fail due to the absence of anomalous images.
Anomaly detection also suffers from ineffective feature extraction due to vague training process.
Our work introduces a novel two-stage anomaly detection algorithm called CONSULT (CONtrastive Self-sUpervised Learning for few-shot Tumor detection). 
The first stage of CONSULT fine-tunes a pre-trained feature extractor specifically for MRI brain images, using a synthetic data generation pipeline to create tumor-like data. 
This process overcomes the lack of anomaly samples and enables the integration of attention mechanisms to focus on anomalous image segments.
The first stage is to overcome the shortcomings of current anomaly detection in extracting features in high-variation data by incorporating Context-Aware Contrastive Learning and Self-supervised Feature Adversarial Learning. 
The second stage of CONSULT uses PatchCore for conventional feature extraction via the fine-tuned weights from the first stage.
To summarize, we propose a self-supervised training scheme for anomaly detection, enhancing model performance and data reliability. 
Furthermore, our proposed contrastive loss, Tritanh Loss, stabilizes learning by offering a unique solution all while enhancing gradient flow. 
Finally, CONSULT achieves superior performance in few-shot brain tumor detection, demonstrating significant improvements over PatchCore by 9.4\%, 12.9\%, 10.2\%, and 6.0\% for 2, 4, 6, and 8 shots, respectively, while training exclusively on healthy images.

\end{abstract}

\section{Introduction}
\label{sec:introduction}
\begin{figure}[t]
    \centering
    \begin{subfigure}[b]{0.24\linewidth}
        \includegraphics[width=\linewidth]{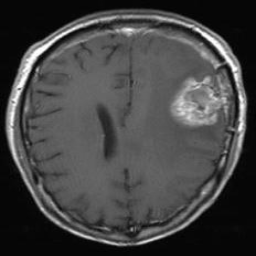}
    \end{subfigure}
    \hfill 
    \begin{subfigure}[b]{0.24\linewidth}
        \includegraphics[width=\linewidth]{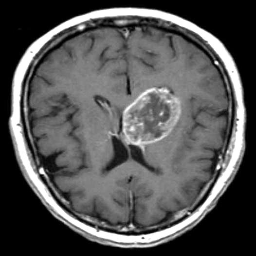}
    \end{subfigure}
    \hfill
    \begin{subfigure}[b]{0.24\linewidth}
        \includegraphics[width=\linewidth]{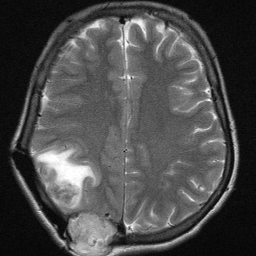}
    \end{subfigure}
    \hfill
    \begin{subfigure}[b]{0.24\linewidth}
        \includegraphics[width=\linewidth]{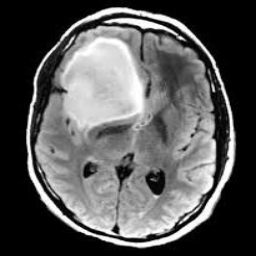}
    \end{subfigure}

    \begin{subfigure}[b]{0.24\linewidth}
        \includegraphics[width=\linewidth]{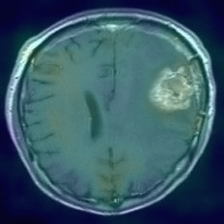}
    \end{subfigure}
    \hfill
    \begin{subfigure}[b]{0.24\linewidth}
        \includegraphics[width=\linewidth]{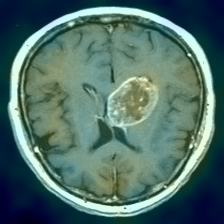}
    \end{subfigure}
    \hfill
    \begin{subfigure}[b]{0.24\linewidth}
        \includegraphics[width=\linewidth]{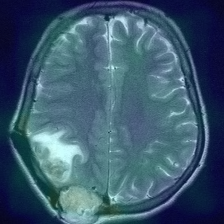}
    \end{subfigure}
    \hfill
    \begin{subfigure}[b]{0.24\linewidth}
        \includegraphics[width=\linewidth]{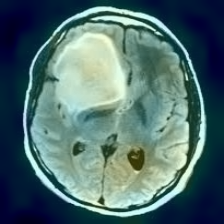}
    \end{subfigure}

    \begin{subfigure}[b]{0.24\linewidth}
        \includegraphics[width=\linewidth]{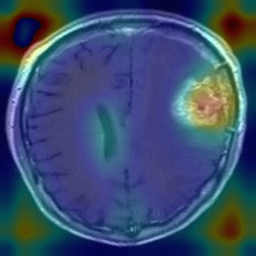}
    \end{subfigure}
    \hfill
    \begin{subfigure}[b]{0.24\linewidth}
        \includegraphics[width=\linewidth]{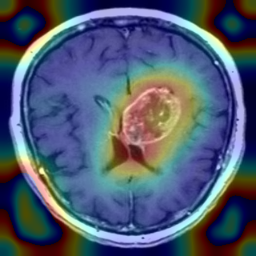}
    \end{subfigure}
    \hfill
    \begin{subfigure}[b]{0.24\linewidth}
        \includegraphics[width=\linewidth]{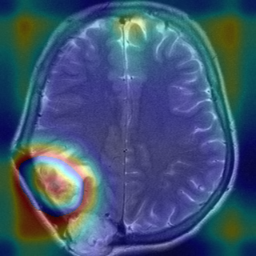}
    \end{subfigure}
    \hfill
    \begin{subfigure}[b]{0.24\linewidth}
        \includegraphics[width=\linewidth]{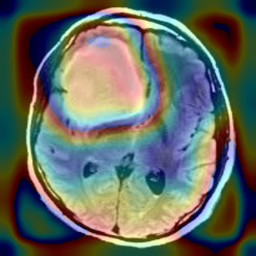}
    \end{subfigure}

    \caption{Heatmap Comparison of Unhealthy Images. Different columns show unhealthy MRI images of different positions, sizes, and textures. The first row is the raw image. The second row is the heatmap predicted with PatchCore. The third raw is the heatmap predicted with the proposed method.}
    \label{fig:comparison-patchcore-CONSULT}
\end{figure}

Brain tumors are one of the most deadly cancers in modern medicine \cite{sung2021global}. Identifying brain tumors at an early stage enables medical professionals to implement prompt interventions and explore treatment methods that are less invasive, significantly improving patient survival rates and reducing negative health impacts.
In contemporary medical diagnostics, Magnetic Resonance Imaging (MRI) has emerged as a cornerstone for the non-invasive detection of brain tumors, leveraging its unparalleled capacity to provide detailed soft tissue contrast \cite{gawande2017brain}. 

Nevertheless, the expense of MRI procedures is high, which is largely attributed to their operational and maintenance expenses. Interpreting MRI images requires the involvement of highly trained and skilled medical professionals, whose salaries further boost the cost \cite{goel2021economic}. 
Moreover, the interpretation of MRIs remains a time-consuming task, posing a challenge for both the energy and expertise of the radiologists. 
Those obstacles have spurred interest in applying deep learning techniques to assist tumor recognition. \cite{huang2020unet, cao2021swinunet, yan2022after, litjens2017survey, hosny2018artificial}
Data scarcity is an unavoidable curse in the field of medical imaging.
This is because collecting medical data for commercial use is faced with challenges and restrictions such as privacy and safety concerns.
Unfortunately, deep learning models need a lot of training data to make an accurate prediction.

To address the aforementioned problems, we argue that anomaly detection is a more suitable task for MRI tumor recognition, which offers both convenience and an explainable recognition method for AI-assisted medical treatment. 
This approach requires only normal data for the \textit{training} phase, removing the need for tumorous data for training.
Instead of learning the anomalous regions (brain tumors), anomaly detection pivots the model to learn the normality of the training dataset. 
Implementing anomaly detection in brain tumor detection can accurately identify anomalies that was unseen in the \textit{training} phase.
Therefore, the advantages of implementing anomaly detection into MRI recognition are clear: 
Firstly, obtaining a comprehensive collection of anomalous data is nearly impossible, whereas acquiring healthy training images is considerably easier.  
Secondly, by exclusively utilizing healthy data for training, the necessity for expert annotations is eliminated, thereby significantly reducing costs. 
Lastly, the predictions made by anomaly detection are traceable back to the training samples, making anomaly detection highly trustworthy.

Although anomaly detection presents a promising approach for analyzing brain MRI images, it is still faced with the final challenge of high variability within brain MRI images.
The heterogeneity of human anatomy poses difficulties in attaining consistent effectiveness across different individuals, both pathological and healthy. 
The classical method to address this issue is to increase the training samples. However, this is not a scalable and feasible solution for MRI images.
Furthermore,  current anomaly detection methods struggle to adapt to the medical imaging domain because of the lack of rigid objective function to fine-tune downstream anomaly detection algorithms.
The objective function for downstream anomaly detection algorithm is extremely hard to define due to the lack of training target and expected outcome.

In response to these problems faced by anomaly detection in MRI images, we propose CONSULT, a two-stage anomaly detection algorithm for few-shot MRI image recognition. 
CONSULT fine-tunes the feature extractor to adapt to the MRI image domain with self-supervised learning. 
The fine-tuned feature extractor is able to differentiate normal features from abnormal ones by focusing on the latter, and it also learns the normality features of healthy samples, reducing the number of training samples needed.
The first stage of CONSULT fine-tune feature extractor with healthy images only.
We deploy self-supervised contrastive learning and self-supervised feature adversarial learning for feature extractor fine-tuning.
Furthermore, we incorporate a lightweight attention module into the feature extractor to help the model focus on anomalous patches iteratively. 
Incorporating context-aware attention module was previously impossible to be applied in downstream anomaly detection because downstream anomaly detection does not undergo any training process.
The second stage is a standard downstream anomaly detection training with the fine-tuned backbone from the first stage. 
We choose PatchCore as the main anomaly detection algorithm to benchmark the effectiveness of CONSULT \cite{roth2022towards}. This is because PatchCore exhibits strong feature sampling and robustness.
Besides making accurate predictions with only two training samples, the proposed self-supervised fine-tuning stage outperforms traditional supervised fine-tuning techniques, such as transfer learning.

Our contributions are threefold:
\begin{itemize}
    \item We propose a self-supervised training scheme for few-shot MRI anomaly detection that does not require tumorous data. Specifically, we introduce context-aware contrastive learning to distinguish healthy features from unhealthy features, and a self-supervised feature adversarial learning to understand the normality of healthy features, thus reducing the need for large training samples.
    \item We develop a novel contrastive loss function, Tritanh loss, which provides stability and robust gradient flow. Unlike anchor loss, which has multiple solutions and a constant gradient, Tritanh loss offers a unique solution and an adaptive gradient flow as the loss approaches optimal.
    \item Through extensive experiments, we show that CONSULT achieves superior performance in few-shot brain tumor detection, significantly improving over PatchCore by 9.4\%, 12.9\%, 10.2\%, and 6.0\% for 2, 4, 6, and 8 shots. 
\end{itemize}

The rest of this paper is organized as follows: In Section II,
we introduce the work related to anomaly detection, medical 
image segmentation, and unsupervised domain adaptation. 
In Section III, we present the methodology of CONSULT. In Section IV, extensive
experiments on the proposed method are conducted to validate
the effectiveness of our method. We conclude our work in
Section V.

\section{Previous Work}

In recent times, the field of anomaly detection has garnered significant attention from both industrial and academic communities. Owing to its ability to deliver precise results through unsupervised learning, anomaly detection eliminates the necessity for labeled datasets in the development of dependable models. Moreover, anomaly detection are highly reliable and logical coherent, as the outcomes they produce are explainable through a detailed examination of the model's outputs.
Anomaly detection is generally classified into two taxonomy: reconstruction-based approach and downstream-based approach. Both methods have their pros and cons when performing anomaly detection training and detection.

\subsection{Reconstruction-Based Method}
Early works of reconstruction-based anomaly detection involve the use of a simple autoencoder to reconstruct the input image \cite{wei2021real}. During the training phase, the model is exclusively fed normal images, enabling it to learn how to reconstruct these inputs accurately. Therefore, it is expected to only proficiently reconstruct normal images while struggling with anomalous ones. Consequently, the discrepancy between the input image and its reconstruction can serve as an indicator of anomaly, assessed via a distance function. A predefined threshold is then applied to identify the type and location of anomalies.
GANomaly is the first to introduce GAN architecture into anomaly detection. The loss function for GANomaly is modified with an adversarial loss \cite{akcay2019ganomaly}. Following their previous work, Skip-GANomaly introduces skip connection in the autoencoder for better feature learning \cite{akccay2019skip}. f-AnoGAN is then proposed to perform anomaly detection on medical images \cite{schlegl2019f}. VT-ADL replaces the encoder with a transformer encoder to harness the ability of the transformer to extract more representative features \cite{mishra2021vt}. AnoViT uses a vision transformer as a feature extractor to extract features and directly pass these features into a custom decoder block to reconstruct the input image \cite{lee2022anovit}. RIAD and InTra inpaint the input image during training to force the model to learn the missing features from the neighbors \cite{zavrtanik2021reconstruction, pirnay2022inpainting}. This method effectively solves the problem of overfitting with U-net architecture. MSTUnet uses the same inpainting method to train a Swin Transformer for image reconstruction \cite{jiang2022masked}. Recent work includes using diffusion models to reconstruct the input image \cite{wolleb2022diffusion}. Diffusion models prove to be better generators compared to their counterparts due to their ability to generate realistic, high-resolution images. By visualizing the output of the reconstruction model, the defect-free image can be estimated from the query images. The detection head is tasked to measure the dissimilarity between the query image and the reconstructed image. 

\subsection{Downstream-Based methods}
Unlike methods focused on reconstruction, downstream approaches are more straightforward. These methods utilize a feature extractor to capture normal feature representations, which are then preserved in an auxiliary memory bank. During the inference phase, the same feature extractor is employed to obtain features from the query image and conduct comparisons with the features stored in the memory bank.
FastFlow attempts to move the latent features to a standard normal distribution using normalizing flow \cite{yu2021fastflow}, where any anomalies are considered as out-of-distribution data. BGAD improved upon FastFlow's work by introducing a contrastive learning approach to train the detection head \cite{yao2023explicit}. 
PaDIM extract feature representations from the intermediate layers and model them using a Mixture of Gaussian (MoG), the distance function is calculated using the Mahalanobis distance \cite{defard2021padim}. RegAD builds upon PaDIM's work and introduced spatial transformer network (STN) into the CNN models and retrained the STN to learn how to spatially transform the intermediate features \cite{huang2022registration}. 
PatchCore used coreset sampling algorithm to sample the most representative features from the training image \cite{roth2022towards}. This simple approach effectively preserves the original representation of the normal feature, enabling traceability while reducing the memory requirement. 

However, as the normal representation becomes more complex and diverse, the memory bank size increases, and this will result in memory overloading. 
Therefore, Reverse Distillation propose a teacher (encoder) and student (decoder) network to perform anomaly detection. The student decoder is trained to reconstruct good features from the teacher encoder network during the training phase and regions that cannot be reconstructed efficiently are considered anomalies \cite{deng2022anomaly}. Reverse Distillation++ introduced simplex noise in the training process as pseudo-bad images and trained the model to suppress such noise \cite{li2023rethinking}. SimpleNet also introduced pseudo-bad features by injecting Gaussian noise into the features extracted from the feature extractor \cite{liu2023simplenet}. EfficientAD combines both reconstruction and knowledge distillation methods, achieving SOTA results and millisecond prediction using lightweight models. EfficientAD used two lightweight models as the teacher and student models and a vanilla autoencoder. The teacher model is distilled from a huge backbone, eliminating the need for a huge feature extractor for anomaly detection. Both the student and autoencoder learn to reconstruct the features extracted from the teacher model \cite{batzner2024efficientad}.

Reconstruction approaches are limited by the bandwidth of the reconstruction network. Furthermore, a large amount of images are needed to train a robust reconstruction-based anomaly detection model. 
On the other hand, the downstream approach does not need a lot of training data to detect anomalies as it only needs representative features to exist in the training set.
Yet, they are weak in detecting anomalies for high data variation because the feature extractor lacks robustness in extracting expressive embeddings.  
Given the intricate nature of medical diagnostics, which demands a focus on fine-grained details, the feature extractor for medical imagery must discern subtler nuances. 
The ImageNet pre-trained backbone commonly used by the downstream approach fails to extract meaningful features in medical images.
The downstream approach is compelled to use ImageNet pre-trained weights as the feature extractor because they lack a clear objective function to retrain the neural network. 
Drawing inspiration from recent advancements in downstream-based anomaly detection that generate pseudo-anomalous images for training, we introduce a novel approach that utilizes rigid contrastive learning for the detection of anomalies in MRI images.

\section{Methodology}

\subsection{Overview}
\label{sec:methodology-overview}
\begin{figure*}[htbp]
  \centering
  \includegraphics[width=\textwidth]{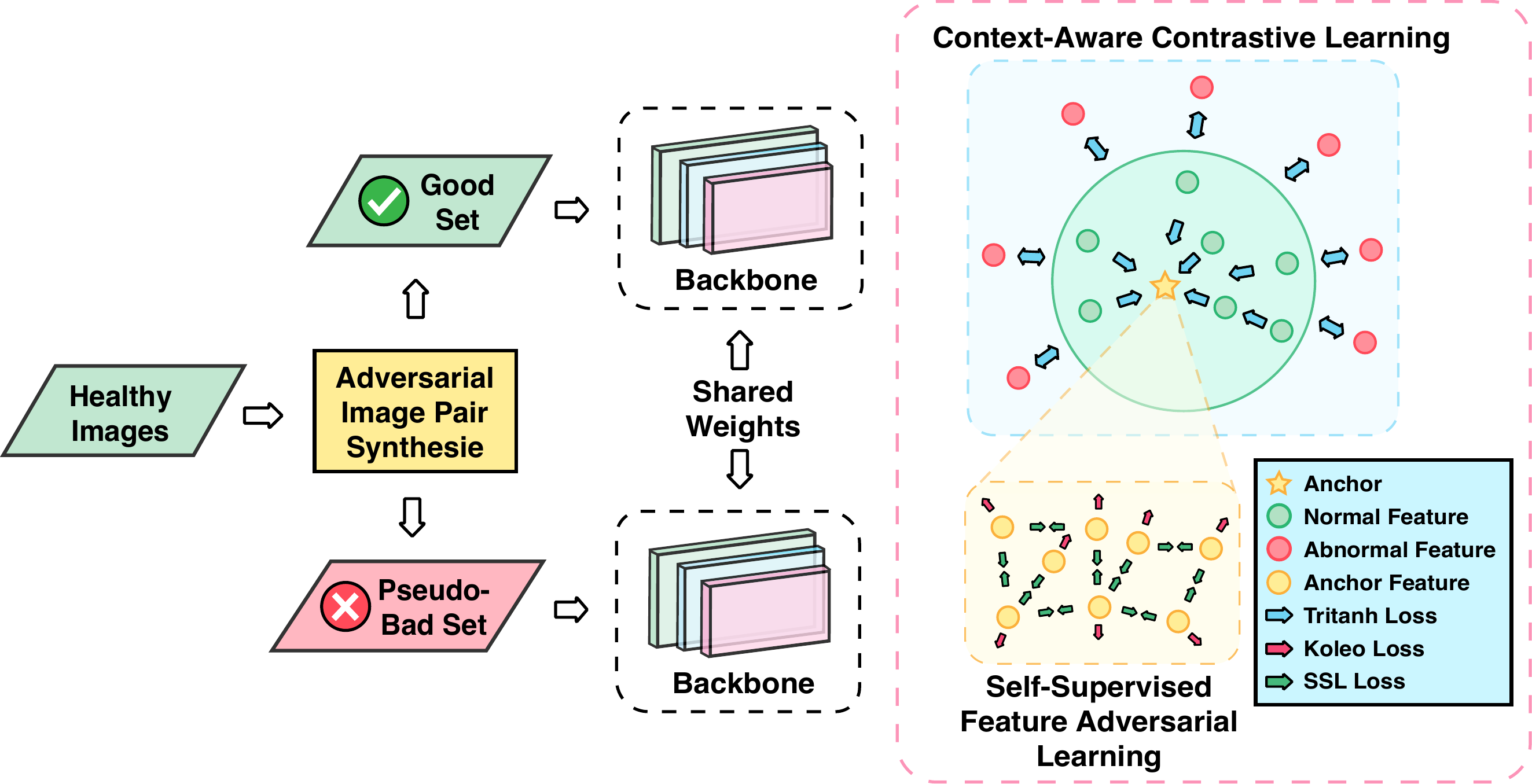}
  \caption{Overall Algorithm for Training Stage. CONSULT takes in $K$ training images and passes them into an augmentation pipeline to generate different representations of healthy and pseudo-unhealthy images. 
  Next, CONSULT trains the backbone through context-aware contrastive learning and self-supervised feature adversarial learning. We use the features extracted from the first, second, and third layers for training. The contrastive learning pipeline helps the model focus on abnormal features by pushing away from the normal features. Furthermore, the normal features are pulled towards each other.
  The self-supervised feature adversarial learning maximizes the self-similarity of the anchor feature by pulling them closer to each other. We also employ a modified KoLeo Loss to prevent catastrophic collapse to ensure the healthy representation is well spread.}
  \label{fig:overall}
\end{figure*}

We introduce CONtrastive Self-sUpervised Learning for few-shot Tumor detection (CONSULT), a 2-stage downstream anomaly detection methodology that achieves extreme-few-shot anomaly detection with high precision. A comprehensive overview of CONSULT is in Figure \ref{fig:overall}.

In the first stage, CONSULT adapts the feature extractor to brain MRI imagery, which enables the extraction of more significant features from healthy brain MRI images. 
To facilitate contrastive learning, we generate a pseudo dataset comprising both tumor-free and synthetic tumorous MRI images. 
For non-tumorous images, we apply image augmentation techniques, such as elastic distortion and random translation to enrich the limited semantic information. 
For the simulation of tumorous images, we employ the Bezier Curve technique to ensure that the artificial defects bear a semantic resemblance to actual brain tumors. 
The goal is to refine the representation of nominal features within the feature space while emphasizing the distinctions between tumorous and non-tumorous features.

To further enhance the efficacy of the feature extractor, we incorporate attention mechanism into our framework. 
Concretely, we deploy an attention module in each layer of the network based on \cite{woo2018cbam}. The attention module prioritizes salient features within the feature map that contribute to minimizing the objective function, making the attention module aware of anomalous features. Previous downstream methods are unable to incorporate context-aware attention modules into anomaly detection because they do not undergo training. 

In the second stage, we replace the feature extractor of PatchCore with the finetuned feature extractor from the first stage \cite{roth2022towards}. The memory bank is built by using coreset sampling algorithm to sample the most representative features from the training samples. 
Our proposed method has a direct improvement on PatchCore and enables PatchCore to make an accurate prediction in low shot scenarios.

PatchCore compares the Euclidian distance between each extracted feature with the sampled points in the memory bank. If the distance is greater than the given threshold, $\tau$, the sample is considered an anomaly. Therefore, we can formulate whether anomaly exists, $A$ in a query image, $x$, when compared with a memory bank, $\mathcal{M}$. 

\begin{equation}
    A = \\
        \left\{
        \begin{array}{ll}
            0 & \quad max(\lvert x(i, j) - \mathcal{M} \rvert ^ 2) \leq \tau \\
            1 & \quad max(\lvert x(i, j) - \mathcal{M} \rvert ^ 2) > \tau
        \end{array}
    \right.
\end{equation}

 It is worth noting that previous studies typically employ over 15 samples for training. In contrast, CONSULT only leverages 2 to 8 shots, the experimental results are detailed in Section \ref{sec:result}. 

\subsection{Adversarial Image Pair Synthesise}

Since anomaly detection models are trained exclusively on normal data, establishing a clear and rigid objective function presents a significant challenge. 
This causes downstream anomaly detection to rely on employing the pre-trained model without any knowledge of the target domain.
This often leads to suboptimal performance and data hunger, especially with complex images with high variability such as brain MRIs. 
Hence, it is imperative to fine-tune the feature extractor using data from the target domain. 
Consequently, the current downstream anomaly detection algorithm requires a substantial amount of healthy images to train a robust anomaly detection model. 
Additionally, it is impractical to amass a comprehensive collection of all possible variations of brain MRI images.

To address these issues, we propose adversarial image pair synthesize, which generates both normal (healthy) and abnormal (tumor-like) images in the first stage of CONSULT. These pairs, denoted as $\mathcal{D}_{train}$, are derived from a limited dataset, $\mathcal{D}_{few}$, which comprises solely of normal images. The $\mathcal{D}_{few}$ dataset is utilized in stage 2 to construct the memory bank for anomaly detection, rather than employing $\mathcal{D}_{train}$.
To preserve the natural attributes of the brain image, we employ a series of data augmentation techniques on $\mathcal{D}_{few}$ to create a set of pseudo-normal images, $\hat{\mathcal{D}_{G}}$. These images are generated using elastic and grid distortion to simulate the brain's internal movement \cite{zhang2023carvemix, garcea2023data}. We also perform a horizontal flip on these pseudo-normal images to reflect the inherent symmetry of the human brain. Finally, we use random contrast adjustment to simulate different levels of contrast and brightness. Consequently, the comprehensive set of normal training images, $\mathcal{D}_{G}$, includes both the augmented images $\hat{\mathcal{D}_{G}}$ and the original non-tumorous images from $\mathcal{D}_{few}$ (initially referred to as $\tilde{\mathcal{D}_{G}}$ to avoid confusion).

Because most brain tumors are convex, circular, and irregularly shaped, it is essential to consider all of these three criteria when attempting to simulate pseudo-defective images. Therefore, we randomly select an image from $\mathcal{D}_{G}$ to generate the pseudo-defective images by inpainting an elliptic discrepancy in the brain image. We denote the pseudo anomalous images set as $\mathcal{D}_{B}$. Inspired by the Bezier Curve properties, we use the Bezier Curve formula to generate these random elliptic defects. First, the rough location of the brain is located by using simple image morphology and thresholding. This is to refine the position of the elliptic defects such that it is not generated outside the brain. Next, we randomly pick several points within the brain and used them as the fixed points in the Bezier Curve. These fixed points are certain to be convex within the convex hull. Finally, we just fill in the convex hull with random Gaussian Noise and inpaint it into the normal image. Our algorithm allows the generation of elliptic defects with different shapes and complexity. By increasing the number of fixed points $N_{C}$ in the Bezier Curve, we increase the complexity of the generated spot. To introduce some degree of randomness in the Bezier Curve, we can adjust the smoothness of the curve about the fixed points with $E$. In this paper, we set $N_{C}=5$ and $E=0.05$. Algorithm \ref{algo:defect-gen} shows the algorithm to generate these random spots.

\begin{algorithm}
    \caption{Algorithm to Generate Elliptic Defects}
    \label{algo:defect-gen}
    \begin{algorithmic}[1]
        \State \textbf{Input:} $shape$, $N\_C$, $E$
        \State \textbf{Output:} Defect image $img$
        
        \State $\mu \gets \text{randint}(0, 255)$ 
        \State $\sigma \gets \text{uniform}(10, 20)$ 
        
        \State $points \gets \text{GetRandomPoints}(N\_C)$
        \State $spot \gets \text{GetBezierCurve}(points, E)$
        
        \State $label \gets$ zero array with dimensions of $shape$
        \State $img \gets$ zero array with dimensions of $shape$
        \State FillPoly($label$, $spot$, $1$)
        
        \State $noise \gets \mathcal{N}(\mu=128,\sigma=128, shape=(15, 15))$
        \State $noise \gets$ Resize($noise$, $shape$)
        
        \State $img\{label = 1\} \gets noise\{label = 1\}$
        
        \State \textbf{return} $img$
    \end{algorithmic}
\end{algorithm}

\subsection{Context-Aware Contrastive Learning for MRI Anomaly Detection}

The image pair synthesis pipeline creates a set of training image pairs consisting of healthy and pseudo-unhealthy images. With these synthesized image pairs, we fine-tune the feature extractor to enable the model to better discriminate between good and bad images. 
We propose context-aware contrastive learning by incorporating a lightweight attention module into the feature extractor, focusing on the anomalous region. 

Contrastive learning is to maximize the similarity of the normal features and better discriminate anomalous features from normal features. Concretely, it maximizes $\lvert x(i, j) - \mathcal{M} \rvert ^ 2$ in the anomalous region and vice versa, where $\mathcal{M}$ is the normal features. First, we select an image from $\tilde{\mathcal{D}_{G}}$ as the anchor image. We pull $\mathcal{D}_{G}$ (excluding the anchor image) closer to the anchor image and push $\mathcal{D}_{B}$ away from the anchor image. 

Next, the attention module learns to pay more attention to anomalous regions and facilitates normal feature generalizability through contrastive learning. Our first approach uses anchor loss, a popular contrastive loss. Anchor loss perfectly fits with anomaly detection problems by minimizing the distance between normal features and maximizing interclass distance. Equation \ref{eq:anchor-loss} shows the formula for anchor loss.

\begin{figure*}[htbp]
    \centering
    \includegraphics[width=\textwidth]{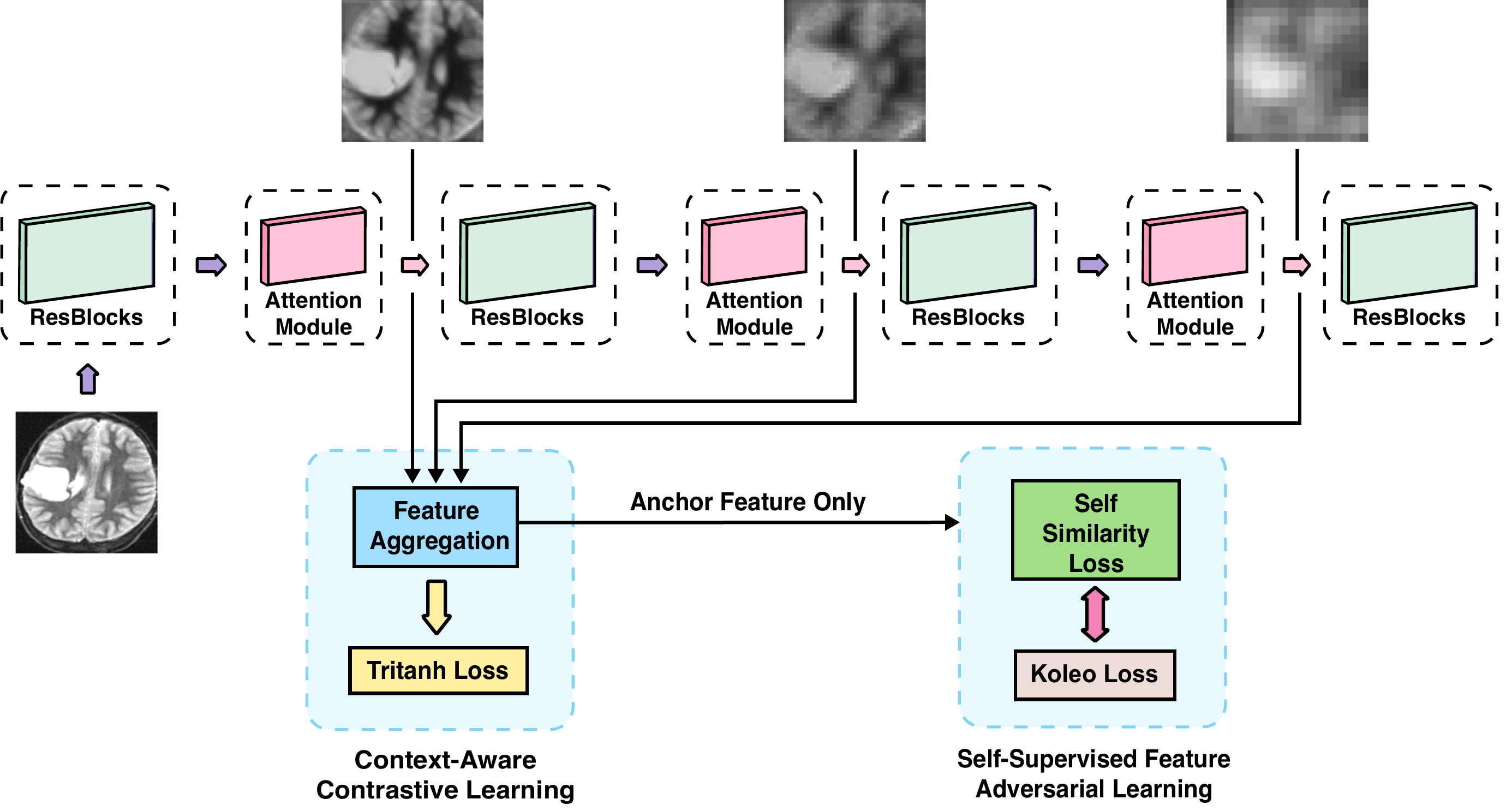}
    \caption{Architecture of the CONSULT Model. 
    CONSULT uses a ResNet backbone as the feature extractor.
    We incorporate context-aware attention modules after each ResBlocks and replace the activation function with Leaky ReLU.
    We choose the first, second, and third feature maps to perform context-aware contrastive learning. Each layer feature is aggregated through resize and concatenation.
    The context-aware contrastive learning trains the attention module to focus on anomalous regions iteratively, enhancing the robustness of feature extraction. An illustrative attention mechanism visualization is provided, showcasing the model's focus on real anomalous regions within MRI images, derived by averaging the first dimension of the attention maps.
    On the other hand, the features of the anchor image are passed to the self-supervised feature adversarial learning to maximize the similarity between the normal features while maintaining representative features. 
    }    
    \label{fig:backbone}
\end{figure*}

\begin{equation}
    \label{eq:anchor-loss}
    L_{anchor} = \max (0, \alpha_0 \cdot d_{\text{pull}} - \alpha_1 \cdot d_{\text{push}} + m )
\end{equation}

However, anchor loss suffers from slow convergence and has multiple solutions, destabilizing training. Inspired by the hyperbolic tangent function, we propose a novel loss function, termed $\text{Tritanh}$ loss, that is more stable and has a more robust gradient flow. Equation \ref{eq:tanh_loss} show the proposed $\text{Tritanh}$ loss.

\begin{equation}
    L_{Tritanh} = \dfrac{e^{\lambda_0 \cdot d_{\text{pull}}} - e^{\lambda_1 \cdot d_{\text{push}}} + m_0}{e^{\lambda_0 \cdot d_{\text{pull}}} + e^{\lambda_1 \cdot d_{\text{push}}} + m_1}
    \label{eq:tanh_loss}
\end{equation}

\begin{figure}[b]
    \centering
    \includegraphics[width=\linewidth]{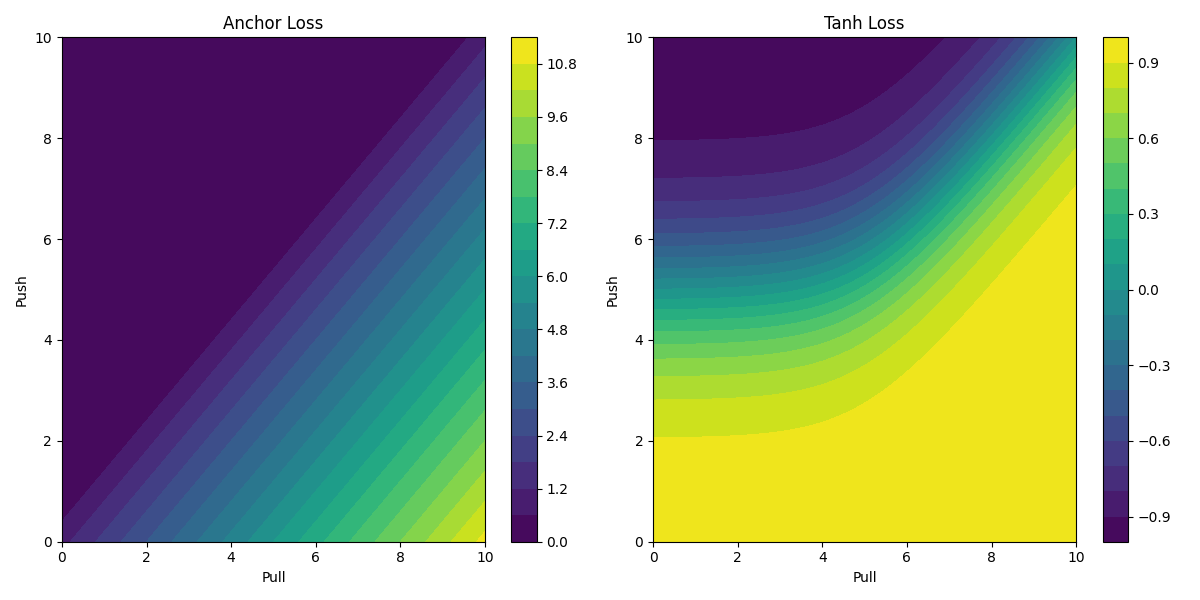}
    \caption{Comparison between Anchor Loss and Tritanh Loss}
    \label{fig:compare-loss}
\end{figure}

$\text{Tritanh}$ loss only has a unique solution where the pulling force, $d_{\text{pull}}$ approaches zero, and the pushing force, $d_{\text{push}}$ approaches infinity. 
Furthermore, we can adjust the gradient of the loss function by changing the scaling factors, $\lambda_0$, and $\lambda_1$ for $d_{\text{pull}}$ and $d_{\text{push}}$ respectively. 
As $\lambda_0 > \lambda_1$, $\text{Tritanh}$ loss prioritizes the pulling force more than the pushing force. We also add two margins, $m_0$ and $m_1$ to regularize the $\text{Tritanh}$ loss. 
By changing these two margins, the gradient of the loss function can be scaled more robustly while being bounded within a controlled value. 
Figure \ref{fig:compare-loss} compares different contrastive loss functions and the proposed contrastive loss function.

As the contrastive loss function minimizes, the attention module iteratively highlights the anomalous regions. 
Figure \ref{fig:backbone} shows the feature maps generated in different layers of the feature extractor. 
We also change the default activation function from ReLU to Leaky ReLU. 
This is because Leaky ReLu allows small negative value, rather than zero gradient flow when a negative value is computed.

\subsection{Self-Supervised Feature Adversarial Learning}

Contrastive loss alone significantly improves the feature extractor's capability to differentiate unhealthy features from healthy features. 
However, it falls short of addressing the problem of high data variation. We propose a self-supervised feature adversarial learning to maximize the intraclass similarity while preventing catastrophic collapse.

We use self-similarity loss to maximize the similarity between healthy image features to further enhance the model generalizability on normal features. We compute the self-similarity loss from the anchor image alone using a simple $L2$ loss. 
Self-similarity loss will make the convex hull of the healthy image features to be more compact. The proposed self-similarity loss is shown in Equation \ref{eq:sim-loss}.

\begin{equation}
    L_{SSL} = \dfrac{1}{N(N-1)} \sum_{i=0}^{N-1} \sum_{j=0}^{N-1} \mathop{\mathbb{I}}(i \neq j) d(P_i, P_j)
    \label{eq:sim-loss}
\end{equation}

For every image patch, $P_k$, we compute the mean of the distance measurement, $d(\dot, \dot)$. 
However, self-similarity loss may cause the feature representation of the anchor image to collapse to a singular point and the extracted features become moot. 

To counter the aforementioned issue, we introduce a regularizer to combat self-similarity loss. 
KoLeo loss preserves the boundaries of the representation while increasing the coverage of respective classes, thus is capable of preventing catastrophic collapse caused by self-similarity loss \cite{sablayrolles2018spreading}. 
The modified KoLeo loss is shown in Equation \ref{eq:koleo-loss}.

\begin{equation}
    L_{KoLeo} = \sum_{x_i \in \mathcal{D}_{train}} - \dfrac{1}{n} \sum_{i=0}^{n-1} \log{|f_\theta (x_i)|}
    \label{eq:koleo-loss}
\end{equation}

Integrating self-similarity loss with KoLeo loss, we have the Feature Adversarial Loss:

\begin{equation}
    L_{SFA} = \gamma_1 L_{SSL} + \gamma_2 L_{KoLeo}
    \label{eq:feature-sdversarial-loss}
\end{equation}

The $\gamma_1$ and $\gamma_2$ are hyperparameters that are empirically set to 0.01. The reduction of self-similarity loss decreases the distance among feature points, consequently diminishing the requisite number of points for representation during coreset sampling. Conversely, KoLeo loss increases the entropy, which serves dual purposes within the feature adversarial learning: firstly, to mitigate the effects of self-similarity loss, thereby averting catastrophic collapse; and secondly, to ensure a more uniform distribution of feature representation.

Therefore, the final loss function for the proposed method is shown in Equation \ref{eq:loss}. 

\begin{equation}
    L = L_{Tritanh} + L_{SFA}
    \label{eq:loss}
\end{equation}

\subsection{Anomaly Detection Head}

Instead of using a pretrained ImageNet feature extractor to obtain the healthy image feature $\mathcal{M}$, we use the fine tuned feature extractor from the first training stage to acquire more representative features. To maintain fairness comparison, build $\mathcal{M}$ only by using $\mathcal{D}_{few}$. Next, we use the coreset sampling method to sample the most representative feature points from $\mathcal{M}$ to create PatchCore memory bank, $\mathcal{M}_{C}^*$. To obtain $\mathcal{M}_{C}^*$, we would need to solve the following minimax problem. 

\begin{equation}
    \mathcal{M}_{C}^* = \argmin_{\mathcal{M}_{C} \subset \mathcal{M}} \max_{m \in \mathcal{M}} \max_{n \in \mathcal{M}_{C}} |m - n|_{2} 
    \label{eq:minmax}
\end{equation}

During inference, we compare the $L2$ distance between each feature extracted from the query image and the memory bank, $\mathcal{M}_{C}^*$. Anomalous image patches are unseen in the memory bank, hence have a greater distance compared to normal features. To increase the robustness of anomaly score, a scaling factor is introduced to reweight the anomaly score by taking in account the nearest neighbours of the closest features in the memory bank.

\section{Experiments}

\subsection{Results}
\label{sec:result}

We choose a public brain MRI images to benchmark CONSULT with previous anomaly detection algorithms \cite{braindataset_chakrabarty}. We only use healthy image for training and move all tumorous images to the validation dataset. Next, we sample 2 ($K2$) to 8 shots ($K8$) of training data from the healthy training set for training CONSULT and discard the rest. All image size are resized to 256 $\times$ 256 pixels. CONSULT is trained with 50 epochs in the first training stage. The learning rate is set to $1e-4$ and we use adam optimizer. 
In the second stage, the coreset sampling ratio is $10\%$ and $K$ nearest neighbor is $9$. 

CONSULT is able to outperform current state-of-the-art anomaly detection algorithms in a low shot scenario. CONSULT shows consistent improvement when more training images are used. PaDIM and PatchCore are two most common downstream anomaly detection algorithms. We also compare CONSULT with reconstruction based anomaly detection such as f-AnoGAN, GANormaly, and EfficientAD. To train a robust model for f-AnoGAN and GANormaly, we increase the training epoch to $100$. The settings for PaDIM and EfficientAD follow the original implementation \cite{defard2021padim,batzner2024efficientad}. The results are shown in Table \ref{tab:primary-res}. 

\begin{table}
\caption{Image AUROC for Few Shot Brain Tumor Detection}
\label{tab:primary-res}
\centering
    \begin{tabular}{c|cccc}
    \textbf{Method} & \textbf{K2} & \textbf{K4} & \textbf{K6} & \textbf{K8} \\
    \hline
    f-AnoGAN  & 0.7291  & 0.8119 & 0.8251  & 0.7751 \\
    GANormaly  & 0.8051  & 0.7669 & 0.7727  & 0.7817 \\
    PaDIM & 0.6151 & 0.7650 & 0.7732 & 0.7710  \\
    EfficientAD & 0.7171 & 0.7554 & 0.7683 & 0.7438  \\
    PatchCore & 0.7727 & 0.7554 & 0.7742 & 0.8138  \\
    Ours & \textbf{0.8663} & \textbf{0.8841} & \textbf{0.8758} & \textbf{0.8738}  \\ 
    \end{tabular}
\end{table}

CONSULT is directly comparable to PatchCore as it uses the anomaly detection head from PatchCore without any changes. This outcome underscores the efficiency of CONSULT in leveraging minimal data for effective learning.

\subsection{Discussion}

The reconstruction-based anomaly detection trains the model to reconstruct the semantic information of healthy samples. However, it need to have high training epoch and a plethora of training samples. It is also worth noting that reconstruction-based anomaly detection offer robustness, but they tend to be less stable and reliable compared to downstream approaches. As the number of training samples increases, we report a lower AUROC for reconstruction approaches. A deeper examination of the reconstructed pictures reveals that, in comparison to the 2 shot experiment, the reconstructed image quality for the 8 shot experiment is poorer. This is because the model is confused due to the increased number of good representations. Downstream approaches lack the robustness inherent in reconstruction-based approaches. This is because the weights used in the downstream approaches are not fine-tuned to the target domain dataset. 

This discrepancy between downstream-based anomaly detection and reconstruction-based approaches catalyzes the development of CONSULT. CONSULT aims to amalgamate the strengths of both downstream and reconstruction approaches. By leveraging unsupervised contrastive learning for backbone fine-tuning, we have demonstrated significant improvements in PatchCore. Specifically, 9.4\%, 12.9\%, 10.2\%, and 6.0\% for 2, 4, 6, and 8 shots respectively. The improvement for our solution primarily comes from the pretraining stage. The objective of the pretraining stage respect the way anomaly detection is being done. Therefore, it can extract more compact features compared to ImageNet, improving the robustness of the feature extractor. In section \ref{sec:ablation1}, we show our pretraining stage proves to be more effective in extracting features for anomaly detection compared to a supervised training method such as classification and segmentation.

Traditional anchor loss, a typical loss function for contrastive learning is unbounded and can result in multiple solutions. Furthermore, choosing the optimal parameters for anchor loss is extremely hard as it will directly affect the training objective. We proposed a novel loss function that is bounded and more robust compared to anchor loss. $\text{Tritanh}$ loss only has a single solution, and the parameters for $\text{Tritanh}$ loss have less effect on the training objectives. Besides, the gradient for $\text{Tritanh}$ loss is steeper compared to anchor loss. The implementation of $\text{Tritanh}$ loss significantly improves upon anchor loss both in robustness and training stability. Next, we introduce other loss functions and regularizers to aid and stabilize the training process. Self-similarity loss maximizes the similarity of healthy features, effectively reducing the need for large training data. It is also coherent with the way PatchCore evaluates anomalies. However, self-similarity loss is very unstable and will result in catastrophic collapse to a singular point as the training epoch progresses. To counter this problem, we introduce a KoLeo Loss as a regularizer for self-similarity loss. We modify the input for KoLeo Loss to be the final feature output from the backbone. Doing so allows the final feature to be more well spread, opposing the self-similarity loss. In section \ref{sec:ablation2}, we compare the effectiveness of the proposed loss functions on CONSULT.

The attention module iteratively helps the model "focus" on regions that are considered anomalous. As shown in Figure \ref{fig:backbone}, the feature maps predicted from each layer focus on the brain tumor region of the query image. Next, we change the activation function of the feature extractor from ReLu to Leaky ReLu for better gradient flow. These improvements help the model to extract a more robust feature compared to a standard ResNet backbone. In section \ref{sec:ablation3}, we compare different feature extractors and the modifications we did to them.


\subsection{Ablation Study}
\subsubsection{Evaluation of CONSULT for Domain Adaptation}
\label{sec:ablation1}

To empirically demonstrate the effectiveness of CONSULT in backbone fine-tuning, we conducted a comprehensive benchmarking experiment by comparing the performance of CONSULT against conventional fine-tuning methods derived from classification and segmentation tasks. The aim is to provide a clear and objective assessment of how CONSULT enhances the backbone's adaptability and accuracy in anomaly detection within medical imaging. Additionally, to ensure fairness and relevance in our comparisons, the fine-tuned backbone from these experiments was integrated into PatchCore. This integration allowed us to assess how effectively the CONSULT-enhanced backbone performs in real-world anomaly detection scenarios.

The dataset used for training the classification and segmentation model is the testing dataset used for benchmarking the anomaly detection algorithm. In other words, the backbone is fully aware of the defects' semantic information, making this comparison significantly unfair to CONSULT. We use ResNet18 as the backbone for the classification model and a head with an output size of 2. For the segmentation model, we use U-Net++ architecture and ResNet18 as its encoder backbone. The classification model achieved a validation accuracy score of 1.0 while the segmentation model achieved a validation IoU of 0.99. Although this benchmarking is unfair to us, CONSULT proved to be an effective unsupervised learning method to fine-tune a backbone for better feature extraction compared to the conventional fully supervised method.

\begin{table}
    \centering
      \caption{Ablation Study: Domain Adaptation}
      \label{tab:ablation-da-res}
      \begin{tabular}{ c | c | c c c c }
        \toprule
        Model & Anno. & K2 & K4 & K6 & K8 \\
        \hline
        \midrule

        CLS & \cmark  & 0.7376  & 0.7074 & 0.7739  & 0.7411 \\
        SEG & \cmark  & 0.8273  & 0.6904 & 0.6887  & 0.6173 \\
        Our & \xmark & \textbf{0.8663}  & \textbf{0.8841} & \textbf{0.8758} & \textbf{0.8686} \\

        \bottomrule
      \end{tabular}
  \end{table}

\subsubsection{Evaluation of Loss Function}
\label{sec:ablation2}
We improve upon anchor loss by addressing several pain points: 1) boundlessness, 2) multiple solutions, and 3) weak gradient flow. The proposed contrastive loss function, inspired by the Euler form of the hyperbolic tangent function, is termed $\text{Tritanh}$ loss.

$\text{Tritanh}$ loss is bounded from -1 to 2 with optimal solution reached will reach when $d_{pull} \longrightarrow 0$ and $d_{push} \longrightarrow \infty$. This ensures that the computed loss does not destabilize the training process.
In contrast to anchor loss which will result in multiple solutions as long as the pushing force is greater than the pulling force by a margin, $\text{Tritanh}$ loss only has a single unique solution. Multiple minima can make the training process more challenging because the optimization algorithm may get stuck in local minima rather than finding the global minimum. 
Additionally, due to the exponential gradient flow, $\text{Tritanh}$ loss gradually slows down as it approaches the minimum, making a near-optimal solution ($d_{pull} \approx 0$ and $d_{push} \approx \infty$) almost equal to the true optimal solution. In practice, achieving the true optimal solution is not necessary; an approximation of it is sufficient for effective performance.
Furthermore, the gradient flow when $d_{pull} \approx d_{push}$ is stepper compared to anchor loss, which only has a constant gradient flow helping the model to converge faster.
Table \ref{tab:ablation-loss-res} compares the result obtained from different loss functions.

\begin{table}
    \centering
      \caption{Ablation Study: Loss Function}
      \label{tab:ablation-loss-res}
      \begin{tabular}{c c c | c c c c }
        \toprule
        Loss & + SSL & + KoLeo & K2 & K4 & K6 & K8 \\
        \hline
        \midrule
        Anchor & \xmark  & \xmark & 0.7457 & 0.7527 & 0.6833 & 0.6912 \\
        Tritanh   & \xmark  & \xmark & 0.791 & 0.8314 & 0.8415 & 0.8305 \\
        Tritanh   & \xmark  & \cmark & 0.7820 & 0.8313 & 0.8486 & 0.8254 \\
        Tritanh   & \cmark  & \xmark & 0.8508 & 0.8531 & 0.8558 & 0.8451 \\
        Tritanh   & \cmark  & \cmark & \textbf{0.8663} & \textbf{0.8841} & \textbf{0.8758} & \textbf{0.8738}  \\ 

        \bottomrule
      \end{tabular}
  \end{table}



\subsubsection{Evaluation of Feature Extractor}
\label{sec:ablation3}
In this section, we compare the modification we have done on the feature extractor. We compare two backbones, ResNet18 and ResNet50. There are two modifications to the traditional backbone that enable us to obtain a richer feature from the query images. The first is the introduction of attention modules after the ResBlocks. The attention modules will help to focus on the defective areas of the query images and it can be seen highlighting defective regions when performing inference. Next, we change the activation function from ReLu to Leaky ReLu to enable negative values in the extracted features, further enriching the feature extraction process.

\begin{table}
    \label{tab:model-res}
    \centering
      \caption{Ablation Study: Model Selection}
      \begin{tabular}{ c c c | c c c c}
        \toprule
        Model & Attn. & Act. & K2 & K4 & K6 & K8 \\
        \hline
        \midrule
        ResNet18 & None  & ReLu        & 0.8296 & 0.8502 & 0.8538 & 0.8512 \\
        ResNet18 & Attention  & ReLu        & 0.8559 & 0.8572 & 0.8444 & 0.8591 \\
        ResNet18 & None  & Leaky  & 0.8587 & 0.8541 & 0.8450 & 0.8588 \\
        ResNet18 & Attention  & Leaky  & \textbf{0.8663} & \textbf{0.8841} & \textbf{0.8758} & \textbf{0.8686} \\

        ResNet50 & None  & ReLu        & 0.8063 & 0.8477 & 0.8518 & 0.8539 \\
        ResNet50 & Attention  & ReLu        & 0.7885 & 0.8507 & 0.8566 & \textbf{0.8636} \\
        ResNet50 & None  & Leaky  & \textbf{0.8557} & 0.8531 & 0.8569 & 0.8558 \\
        ResNet50 & Attention  & Leaky  & 0.8500 & \textbf{0.8558} & \textbf{0.8631} & 0.8408 \\
        \bottomrule
      \end{tabular}
  \end{table}

For both ResNet18 and ResNet50, the modifications to the feature extractor show improvement in extracting more robust features by introducing attention modules and increasing the feature activation function.

\section{Conclusion}

In this paper, we proposed CONSULT for few-shot anomaly detection in MRI images. Anomaly detection is effective for medical images because it relies solely on healthy images for training, eliminating the need for labeled data. CONSULT addresses this by combining the strengths of downstream and reconstruction-based anomaly detection algorithms. We introduced a stringent objective function to fine-tune the feature extractor, enhancing robustness while reducing the need for an abundance of training data. Our results show that CONSULT significantly improves PatchCore in MRI disease recognition and outperforms other reconstruction methods. Future work will extend CONSULT to various datasets, including other medical images, traffic anomalies, and industrial datasets.

\section*{Acknowledgment}

This paper is completed during the industrial attachment with ViTrox Corporation Sdn. Bhd.

\printbibliography

@article{yu2021fastflow,
  title={Fastflow: Unsupervised anomaly detection and localization via 2d normalizing flows},
  author={Yu, Jiawei and Zheng, Ye and Wang, Xiang and Li, Wei and Wu, Yushuang and Zhao, Rui and Wu, Liwei},
  journal={arXiv preprint arXiv:2111.07677},
  year={2021}
}

@article{schlegl2019f,
  title={f-AnoGAN: Fast unsupervised anomaly detection with generative adversarial networks},
  author={Schlegl, Thomas and Seeb{\"o}ck, Philipp and Waldstein, Sebastian M and Langs, Georg and Schmidt-Erfurth, Ursula},
  journal={Medical image analysis},
  volume={54},
  pages={30--44},
  year={2019},
  publisher={Elsevier}
}

@article{zavrtanik2021reconstruction,
  title={Reconstruction by inpainting for visual anomaly detection},
  author={Zavrtanik, Vitjan and Kristan, Matej and Sko{\v{c}}aj, Danijel},
  journal={Pattern Recognition},
  volume={112},
  pages={107706},
  year={2021},
  publisher={Elsevier}
}

@inproceedings{deng2022anomaly,
  title={Anomaly detection via reverse distillation from one-class embedding},
  author={Deng, Hanqiu and Li, Xingyu},
  booktitle={Proceedings of the IEEE/CVF Conference on Computer Vision and Pattern Recognition},
  pages={9737--9746},
  year={2022}
}

@article{sablayrolles2018spreading,
  title={Spreading vectors for similarity search},
  author={Sablayrolles, Alexandre and Douze, Matthijs and Schmid, Cordelia and J{\'e}gou, Herv{\'e}},
  journal={arXiv preprint arXiv:1806.03198},
  year={2018}
}

@inproceedings{woo2018cbam,
  title={Cbam: Convolutional block attention module},
  author={Woo, Sanghyun and Park, Jongchan and Lee, Joon-Young and Kweon, In So},
  booktitle={Proceedings of the European conference on computer vision (ECCV)},
  pages={3--19},
  year={2018}
}

@article{hosny2018artificial,
  title={Artificial intelligence in radiology},
  author={Hosny, Ahmed and Parmar, Chintan and Quackenbush, John and Schwartz, Lawrence H and Aerts, Hugo JWL},
  journal={Nature Reviews Cancer},
  volume={18},
  number={8},
  pages={500--510},
  year={2018},
  publisher={Nature Publishing Group}
}

@article{litjens2017survey,
  title={A survey on deep learning in medical image analysis},
  author={Litjens, Geert and Kooi, Thijs and Bejnordi, Babak Ehteshami and Setio, Arnaud Arindra Adiyoso and Ciompi, Francesco and Ghafoorian, Mohsen and Van Der Laak, Jeroen Awm and Van Ginneken, Bram and S{\'a}nchez, Clara I},
  journal={Medical image analysis},
  volume={42},
  pages={60--88},
  year={2017},
  publisher={Elsevier}
}

@article{goel2021economic,
  title={Economic implications of the modern treatment paradigm of glioblastoma: an analysis of global cost estimates and their utility for cost assessment},
  author={Goel, Nicholas J and Bird, Cylaina E and Hicks, William H and Abdullah, Kalil G},
  journal={Journal of Medical Economics},
  volume={24},
  number={1},
  pages={1018--1024},
  year={2021},
  publisher={Taylor \& Francis}
}

@inproceedings{gawande2017brain,
  title={Brain tumor diagnosis using image processing: A survey},
  author={Gawande, Suraj S and Mendre, Vrushali},
  booktitle={2017 2nd IEEE International Conference on Recent Trends in Electronics, Information \& Communication Technology (RTEICT)},
  pages={466--470},
  year={2017},
  organization={IEEE}
}

@article{sung2021global,
  title={Global cancer statistics 2020: GLOBOCAN estimates of incidence and mortality worldwide for 36 cancers in 185 countries},
  author={Sung, Hyuna and Ferlay, Jacques and Siegel, Rebecca L and Laversanne, Mathieu and Soerjomataram, Isabelle and Jemal, Ahmedin and Bray, Freddie},
  journal={CA: a cancer journal for clinicians},
  volume={71},
  number={3},
  pages={209--249},
  year={2021},
  publisher={Wiley Online Library}
}

@inproceedings{huang2020unet,
  title={Unet 3+: A full-scale connected unet for medical image segmentation},
  author={Huang, Huimin and Lin, Lanfen and Tong, Ruofeng and Hu, Hongjie and Zhang, Qiaowei and Iwamoto, Yutaro and Han, Xianhua and Chen, Yen-Wei and Wu, Jian},
  booktitle={ICASSP 2020-2020 IEEE international conference on acoustics, speech and signal processing (ICASSP)},
  pages={1055--1059},
  year={2020},
  organization={IEEE}
}

@inproceedings{batzner2024efficientad,
  title={Efficientad: Accurate visual anomaly detection at millisecond-level latencies},
  author={Batzner, Kilian and Heckler, Lars and K{\"o}nig, Rebecca},
  booktitle={Proceedings of the IEEE/CVF Winter Conference on Applications of Computer Vision},
  pages={128--138},
  year={2024}
}

@inproceedings{liu2023simplenet,
  title={Simplenet: A simple network for image anomaly detection and localization},
  author={Liu, Zhikang and Zhou, Yiming and Xu, Yuansheng and Wang, Zilei},
  booktitle={Proceedings of the IEEE/CVF Conference on Computer Vision and Pattern Recognition},
  pages={20402--20411},
  year={2023}
}

@inproceedings{li2023rethinking,
  title={Rethinking Feature-Based Knowledge Distillation for Face Recognition},
  author={Li, Jingzhi and Guo, Zidong and Li, Hui and Han, Seungju and Baek, Ji-won and Yang, Min and Yang, Ran and Suh, Sungjoo},
  booktitle={Proceedings of the IEEE/CVF Conference on Computer Vision and Pattern Recognition},
  pages={20156--20165},
  year={2023}
}

@inproceedings{huang2022registration,
  title={Registration based few-shot anomaly detection},
  author={Huang, Chaoqin and Guan, Haoyan and Jiang, Aofan and Zhang, Ya and Spratling, Michael and Wang, Yan-Feng},
  booktitle={European Conference on Computer Vision},
  pages={303--319},
  year={2022},
  organization={Springer}
}

@inproceedings{wolleb2022diffusion,
  title={Diffusion models for medical anomaly detection},
  author={Wolleb, Julia and Bieder, Florentin and Sandk{\"u}hler, Robin and Cattin, Philippe C},
  booktitle={International Conference on Medical image computing and computer-assisted intervention},
  pages={35--45},
  year={2022},
  organization={Springer}
}

@article{jiang2022masked,
  title={Masked swin transformer unet for industrial anomaly detection},
  author={Jiang, Jielin and Zhu, Jiale and Bilal, Muhammad and Cui, Yan and Kumar, Neeraj and Dou, Ruihan and Su, Feng and Xu, Xiaolong},
  journal={IEEE Transactions on Industrial Informatics},
  volume={19},
  number={2},
  pages={2200--2209},
  year={2022},
  publisher={IEEE}
}

@inproceedings{pirnay2022inpainting,
  title={Inpainting transformer for anomaly detection},
  author={Pirnay, Jonathan and Chai, Keng},
  booktitle={International Conference on Image Analysis and Processing},
  pages={394--406},
  year={2022},
  organization={Springer}
}

@article{lee2022anovit,
  title={AnoViT: Unsupervised anomaly detection and localization with vision transformer-based encoder-decoder},
  author={Lee, Yunseung and Kang, Pilsung},
  journal={IEEE Access},
  volume={10},
  pages={46717--46724},
  year={2022},
  publisher={IEEE}
}

@inproceedings{mishra2021vt,
  title={VT-ADL: A vision transformer network for image anomaly detection and localization},
  author={Mishra, Pankaj and Verk, Riccardo and Fornasier, Daniele and Piciarelli, Claudio and Foresti, Gian Luca},
  booktitle={2021 IEEE 30th International Symposium on Industrial Electronics (ISIE)},
  pages={01--06},
  year={2021},
  organization={IEEE}
}

@inproceedings{akcay2019ganomaly,
  title={Ganomaly: Semi-supervised anomaly detection via adversarial training},
  author={Akcay, Samet and Atapour-Abarghouei, Amir and Breckon, Toby P},
  booktitle={Computer Vision--ACCV 2018: 14th Asian Conference on Computer Vision, Perth, Australia, December 2--6, 2018, Revised Selected Papers, Part III 14},
  pages={622--637},
  year={2019},
  organization={Springer}
}

@inproceedings{akccay2019skip,
  title={Skip-ganomaly: Skip connected and adversarially trained encoder-decoder anomaly detection},
  author={Ak{\c{c}}ay, Samet and Atapour-Abarghouei, Amir and Breckon, Toby P},
  booktitle={2019 International Joint Conference on Neural Networks (IJCNN)},
  pages={1--8},
  year={2019},
  organization={IEEE}
}

@article{wei2021real,
  title={Real-time implementation of fabric defect detection based on variational automatic encoder with structure similarity},
  author={Wei, Wei and Deng, Dexiang and Zeng, Lin and Zhang, Chen},
  journal={Journal of Real-Time Image Processing},
  volume={18},
  pages={807--823},
  year={2021},
  publisher={Springer}
}

@inproceedings{yao2023explicit,
  title={Explicit Boundary Guided Semi-Push-Pull Contrastive Learning for Supervised Anomaly Detection},
  author={Yao, Xincheng and Li, Ruoqi and Zhang, Jing and Sun, Jun and Zhang, Chongyang},
  booktitle={Proceedings of the IEEE/CVF Conference on Computer Vision and Pattern Recognition},
  pages={24490--24499},
  year={2023}
}

@article{garcea2023data,
  title={Data augmentation for medical imaging: A systematic literature review},
  author={Garcea, Fabio and Serra, Alessio and Lamberti, Fabrizio and Morra, Lia},
  journal={Computers in Biology and Medicine},
  volume={152},
  pages={106391},
  year={2023},
  publisher={Elsevier}
}

@misc{braindataset_chakrabarty,
    author = {Navoneel Chakrabarty},
    title = {Brain MRI Images for Brain Tumor Detection},
    year = {2018},
    note = "{Retrieved Date Retrieved from \url{https://www.kaggle.com/datasets/navoneel/brain-mri-images-for-brain-tumor-detection}}"
}

@misc{cao2021swinunet,
      title={Swin-Unet: Unet-like Pure Transformer for Medical Image Segmentation}, 
      author={Hu Cao and Yueyue Wang and Joy Chen and Dongsheng Jiang and Xiaopeng Zhang and Qi Tian and Manning Wang},
      year={2021},
      eprint={2105.05537},
      archivePrefix={arXiv},
      primaryClass={eess.IV}
}

@inproceedings{yan2022after,
  title={After-unet: Axial fusion transformer unet for medical image segmentation},
  author={Yan, Xiangyi and Tang, Hao and Sun, Shanlin and Ma, Haoyu and Kong, Deying and Xie, Xiaohui},
  booktitle={Proceedings of the IEEE/CVF winter conference on applications of computer vision},
  pages={3971--3981},
  year={2022}
}

@article{zhang2023carvemix,
  title={CarveMix: a simple data augmentation method for brain lesion segmentation},
  author={Zhang, Xinru and Liu, Chenghao and Ou, Ni and Zeng, Xiangzhu and Zhuo, Zhizheng and Duan, Yunyun and Xiong, Xiaoliang and Yu, Yizhou and           Liu, Zhiwen and Liu, Yaou and others},
  journal={NeuroImage},
  pages={120041},
  year={2023},
  publisher={Elsevier}
  }

@inproceedings{defard2021padim,
  title={Padim: a patch distribution modeling framework for anomaly detection and localization},
  author={Defard, Thomas and Setkov, Aleksandr and Loesch, Angelique and Audigier, Romaric},
  booktitle={International Conference on Pattern Recognition},
  pages={475--489},
  year={2021},
  organization={Springer}
}

@inproceedings{roth2022towards,
  title={Towards total recall in industrial anomaly detection},
  author={Roth, Karsten and Pemula, Latha and Zepeda, Joaquin and Sch{\"o}lkopf, Bernhard and Brox, Thomas and Gehler, Peter},
  booktitle={Proceedings of the IEEE/CVF Conference on Computer Vision and Pattern Recognition},
  pages={14318--14328},
  year={2022}
}

\end{document}